\documentclass[11pt,a4paper]{article}
\usepackage[hyperref]{emnlp2018}
\usepackage{times}
\usepackage{latexsym}
\usepackage{amsmath}
\usepackage{graphicx}
\usepackage{amssymb}
\usepackage{url}

\aclfinalcopy % Uncomment this line for the final submission

\title{Bayesian Compression for Natural Language Processing}

\author{Nadezhda Chirkova$\bf{}^{1}$\thanks{$\quad$Equal contribution.} , Ekaterina Lobacheva$\bf{}^{1*}$, Dmitry Vetrov$\bf{}^{1,2}$\\
	${}^1$Samsung-HSE Laboratory, National Research University Higher School of Economics\\
	${}^2$Samsung AI Center\\
	Moscow, Russia\\
	{\tt \{nchirkova,elobacheva,dvetrov\}@hse.ru} }
	
\date{}

\begin{document}
	\maketitle
	\begin{abstract}
	In natural language processing, a lot of the tasks are successfully solved with recurrent neural networks, but such models have a huge number of parameters. The majority of these parameters are often concentrated in the embedding layer, which size grows proportionally to the vocabulary length. We propose a Bayesian sparsification technique for RNNs which allows compressing the RNN dozens or hundreds of times without time-consuming hyperparameters tuning. We also generalize the model for vocabulary sparsification to filter out unnecessary words and compress the RNN even further. We show that the choice of the kept words is interpretable. 
	\end{abstract}

	\section{Introduction}
	Recurrent neural networks (RNNs) are among the most powerful models 
    for natural language processing, speech recognition, question-answering systems
    ~\cite{las,hypernet,trans,qa}. For complex tasks such as machine translation~\cite{trans} 
    modern RNN architectures incorporate a huge number of parameters. To use these models on portable devices with limited memory
    the model compression is desired. 
    
    There are a lot of RNNs compression methods based on specific weight matrix representations~\cite{tjandra,kroneker} or sparsification~\cite{pruning,groupsparseLSTM}. In this paper we focus on RNNs compression via sparsification. 
    One way to sparsify RNN is pruning where the weights with a small absolute value are eliminated from the model. Such methods are heuristic and require time-consuming hyperparameters tuning. 
    There is another group of sparsification techniques based on Bayesian approach. \citeauthor{dmolch}~\shortcite{dmolch} describe a model called SparseVD in which parameters controlling sparsity are tuned automatically during neural network training.  However, this technique was not previously investigated for RNNs. In this paper, we apply Sparse VD to RNNs taking into account the specifics of recurrent network structure (Section~\ref{sec:sparsevd_rnn}). More precisely, we use the insight 
    about using the same sample of weights for all timesteps in the sequence~\cite{gal,BayesianRNN}. This modification makes local reparametrization trick~\cite{kingma,dmolch} not applicable and changes SparseVD training procedure.
    
    In natural language processing tasks 
    the majority of weights in RNNs are often concentrated in the first layer that is connected to the vocabulary, for example in embedding layer. 
    However, for some tasks the most of the words are  unnecessary for accurate predictions. In our model we introduce multiplicative weights for the words to perform vocabulary sparsification  (Section~\ref{sec:mult}). These multiplicative weights are zeroing out during training causing filtering corresponding unnecessary words out of the model.
    It allows to boost RNN sparsification level even further. 
    
    To sum up, our contributions are as follows: (i) we adapt SparseVD to RNNs explaining the specifics of the resulting model and (ii) we generalize this model by introducing multiplicative weights for words to purposefully sparsify the vocabulary.
	Our results show that Sparse Variational Dropout leads to a very high level of sparsity in recurrent models without a significant quality drop. Models with additional vocabulary sparsification boost compression rate on text classification tasks but do not help that much on language modeling tasks. In classification tasks the vocabulary is compressed dozens of times, and the choice of words is interpretable.
	
	\section{Related work}
	
	Reducing RNN size is an important and rapidly developing area of research. There are three research directions: approximation of weight matriсes~\cite{tjandra, kroneker}, reducing the precision of the weights~\cite{itay} and sparsification of the weight matrices~\cite{pruning, groupsparseLSTM}. We focus on the last one.
	The most popular approach here is pruning: the weights of the RNN are cut off on some threshold. \citeauthor{pruning}~\shortcite{pruning} choose threshold using several hyperparameters that control the frequency, the rate and the duration of the weights eliminating. \citeauthor{groupsparseLSTM}~\shortcite{groupsparseLSTM} propose to prune the weights in LSTM by groups corresponding to each neuron, this allows to accelerate forward pass through the network.
	
	Another group of sparsification methods relies on Bayesian neural networks~\cite{dmolch, lognormal, chris}. In Bayesian NNs the weights are treated as random variables, and our desire about sparse weights is expressed in a prior distribution over them. During training, the prior distribution is transformed into the posterior distribution over the weights, used to make predictions on testing phase. \citeauthor{lognormal}~\shortcite{lognormal} and \citeauthor{chris}~\shortcite{chris} also introduce group Bayesian sparsification techniques that allow to eliminate neurons from the model.

The main advantage of the Bayesian sparsification techniques is that they have a small number of hyperparameters compared to pruning-based methods. Also, they lead to a higher sparsity level~\cite{dmolch, lognormal, chris}.
	
	There are several works on Bayesian recurrent neural networks~\cite{gal, BayesianRNN}, but these methods are hard to extend to achieve sparsification. We apply sparse variational dropout to RNNs taking into account its recurrent specifics, including some insights highlighted by \citeauthor{gal}~\shortcite{gal}, \citeauthor{BayesianRNN}~\shortcite{BayesianRNN}.
	
	\section{Proposed method}
	
	\subsection{Notations}
	In the rest of the paper $x = [x_0, \dots, x_T]$ is an input sequence,
	$y$ is a true output and $\hat{y}$ is an output predicted by the RNN
	($y$ and $\hat{y}$ may be single vectors, sequences, etc.), $X, Y$ denotes a training set $\{(x^1, y^1), \dots, (x^N, y^N)\}$.
	All weights of the RNN except biases are denoted by $\omega$, while a single weight (an element of any weight matrix) is denoted by $w_{ij}$. Note that we detach biases and denote them by $B$ because we do not sparsify them.
	
	For definiteness, we will illustrate our model on an example  architecture for the language modeling task, where $y = [x_1, \dots, x_T]$:
	\begin{align}
&	\text{embedding}: \tilde x_t = w^e_{x_t};  \nonumber \\% \in  \mathbb{R}^k$
	& \text{recurrent}: h_{t+1} = \sigma(W^h h_t + W^x \tilde x_{t+1}+b^r); \nonumber\\ 
	&\text{fully-connected}: \hat{y}_t = \text{softmax}(W^d h_t+b^d).\nonumber
	\end{align}

	In this example $\omega = \{W^e, W^x, W^h, W^d \}$ , $B = \{b^r, b^d\}$. However, the model may be directly applied to any recurrent architecture.
	
	\subsection{Sparse variational dropout for RNNs} \label{sec:sparsevd_rnn}
	
	Following \citeauthor{kingma}~\shortcite{kingma}, \citeauthor{dmolch}~\shortcite{dmolch}, we put a fully-factorized log-uniform prior over the weights:
	\[
    p(\omega) = \prod_{w_{ij} \in \omega} p(w_{ij}) , \quad	p(w_{ij}) \propto \frac 1 {|w_{ij}|}
	\]
and approximate 	the posterior
	with a fully factorized normal distribution:%~\cite{dmolch}:
	\[
	q(w|\theta, \sigma) = \prod_{w_{ij} \in \omega} \mathcal{N}\bigl(w_{ij}|\theta_{ij}, \sigma_{ij}^2\bigr).
	\]
	The task of posterior approximation 
	$%\[
	\min_{\theta, \sigma, B} KL(q(\omega|\theta, \sigma)||p(\omega|X, Y, B))
   $ %	\]
  	is equivalent to variational lower bound optimization~\cite{dmolch}:
	\begin{gather} 
	- \sum_{i=1}^N \int q(\omega|\theta, \sigma) 
	%\log p \Bigl(y_i\big|f_h(x_{i,T}, f_h(\dots f_h (x_{i, 1}, h_{i, 0})))\Bigr)
	\log p(y^i | x^i_0, \dots, x^i_T, \omega, B)
	 d \omega + \notag \\
	 + \sum_{w_{ij} \in \omega} KL(q(w_{ij}|\theta_{ij}, \sigma_{ij})||p(w_{ij}))
	%- \sum_{w_{ij} \in \omega} k\biggl(\frac{{\sigma_{ij}}^2}{{\theta_{ij}}^2}\biggr) 
	\rightarrow \min_{\theta, \sigma, B}.
	\label{elbo}
	\end{gather}
	Here the first term, a task-specific loss, is approximated with one sample from $q(\omega|\theta, \sigma)$. The second term is a regularizer that moves posterior closer to prior and induces sparsity. This regularizer can be very closely approximated analytically~\cite{dmolch}:
	\begin{gather}
		\label{KL}
	KL(q(w_{ij}|\theta_{ij}, \sigma_{ij})||p(w_{ij})) \approx k\biggl(\frac{{\sigma^2_{ij}}}{{\theta^2_{ij}}}\biggr),
	\end{gather}
{\small\begin{gather}
	k(\alpha) \approx 0.64 \sigma (1.87 + 1.49\log \alpha)
	- \frac 1 2\log\Bigl(1 + \frac 1 \alpha \Bigr) .
	% + C
 \nonumber
	\end{gather}}
	% C?
	
	To make integral estimation unbiased, sampling from the posterior is performed with the use of reparametrization trick ~\cite{RT}:
	\begin{equation}
	\label{LRT}
	w_{ij} = \theta_{ij} + \sigma_{ij} \epsilon_{ij},  \quad
	\epsilon_{ij} \sim \mathcal{N}(\epsilon_{ij}|0, 1)
	\end{equation}
    
    The important difference of RNNs compared to feed-forward networks consists in sharing the same weight variable between different timesteps. Thus, we should use the same sample of weights for each timestep $t$ while computing the likelihood $p(y^i | x^i_0, \dots, x^i_T, \omega, B)$~\cite{gal, BayesianRNN}. 
    
    \citeauthor{kingma}~\shortcite{kingma}, \citeauthor{dmolch} \shortcite{dmolch} also use local reparametrization trick (LRT) that is sampling preactivation instead of individual weights. For example,
    \[
    (W^x x_t)_i = \sum_{j} \theta^x_{ij} x_{tj} + \epsilon_i 
     \sqrt{\sum_{j} (\sigma^x_{ij})^2 x^2_{tj}}.
    \]
    Tied weight sampling makes LRT not applicable to weight matrices that are used in more than one timestep in the RNN. 
    
    For the hidden-to-hidden matrix $W^h$ the linear combination $(W^h h_t)$ is not normally distributed because $h_t$ depends on $W^h$ from the previous timestep. As a result, the rule about the sum of independent normal distributions with constant coefficients is not applicable. In practice, network with LRT on hidden-to-hidden weights cannot be trained properly. 
    
    For the input-to-hidden matrix $W^x$ the linear combination $(W^x x_t)$ is normally distributed. However, sampling the same $W^x$ for all timesteps and sampling the same noise $\epsilon_i$ for preactivations for all timesteps are not equivalent. The same sample of $W^x$ corresponds to different samples of noise $\epsilon_i$ at different timesteps because of the different $x_t$. Hence theoretically LRT is not applicable here. In practice, networks with LRT on input-to-hidden weights may give the same results and in some experiments, they even converge a little bit faster.
    
    Since the training procedure is effective only with 2D noise tensor, we propose to sample the noise on the weights per mini-batch, not per individual object.
    
    To sum up, the training procedure is as follows. To perform forward pass for a mini-batch, we firstly sample all weights $\omega$ following~\eqref{LRT} and then apply RNN as usual. Then the gradients of~\eqref{elbo} are computed w.r.t $\theta, \log\sigma, B$. 
   
   	During the testing stage, we use the mean weights $\theta$~\cite{dmolch}. Regularizer~\eqref{KL} causes the majority of $\theta$ components approach 0, and the weights are sparsified. More precisely, we eliminate weights with low signal-to-noise ratio $ 
   	\frac {\theta^2_{ij}}{\sigma^2_{ij}} < \tau $~\cite{dmolch}.
	
	\subsection{Multiplicative weights for vocabulary sparsification} \label{sec:mult}
	One of the advantages of Bayesian sparsification
	is an easy generalization for the sparsification of any groups of the weights
	that doesn't complicate the training procedure~\cite{chris}. 
	To do so, one should introduce shared multiplicative weight per each group, and elimination of this multiplicative weight will mean the elimination of the corresponding group.
	In our work we utilize this approach to achieve vocabulary sparsification.
	
	Precisely, we introduce multiplicative probabilistic weights $z\in\mathbb{R}^V$ for words in the vocabulary (here $V$ is the size of the vocabulary). The forward pass with $z$ looks as follows:
	\begin{enumerate}
	    \item sample vector $z^i$ from the current approximation of the posterior for each input sequence $x^i$ from the mini-batch;
	    \item multiply each one-hot encoded token $x^i_t$ from the sequence $x^i$ by $z^i$ (here both $x^i_t$ and $z^i$ are $V$-dimensional);
	    \item continue the forward pass as usual.
	\end{enumerate}
	
	We work with $z$ in the same way as with other weights $W$: we use a log-uniform prior and approximate the posterior with a fully-factorized normal distribution with trainable mean and variance. However, since $z$ is a one-dimensional vector, we can sample it individually for each object in a mini-batch to reduce the variance of the gradients. After training, we prune elements of $z$ with a low signal-to-noise ratio and subsequently, we do not use the corresponding words from the vocabulary and drop columns of weights from the embedding or input-to-hidden weight matrices. 
	
	\section{Experiments}
We perform experiments with LSTM architecture on two types of problems: text classification and language modeling. 
Three models are compared here: baseline model without any regularization, SparseVD model and SparseVD model with multiplicative weights for vocabulary sparsification (SparseVD-Voc).

To measure the sparsity level of our models we calculate the compression rate of individual weights as follows: $|w|/|w\neq0|$. The sparsification of weights may lead not only to the compression but also to the acceleration of RNNs through group sparsity. Hence, we report the number of remaining neurons in all layers: input (vocabulary), embedding and recurrent. To compute this number for vocabulary layer in SparseVD-Voc we use introduced variables $z_v$. For all other layers in SparseVD and SparseVD-Voc, we drop a neuron if all weights connected to this neuron are eliminated. 

We optimize our networks using Adam~\cite{adam}.  Baseline networks overfit for all our tasks, therefore, we present results for them with early stopping.
For all weights that we sparsify, we initialize $\log\sigma$ with -3. We eliminate weights with signal-to-noise  ratio less then $\tau=0.05$.
More details about experiment setup are presented in Appendix~\ref{setup}.

\begin{table*}[ht!]
  \normalsize
  \centering
  \begin{tabular}{clcccc}
Task & Method & Accuracy \% & Compression  & Vocabulary & Neurons $\tilde{x}$ - $h$\\ 
\hline
&\,Original  & 84.1 & 1x & 20000 & $300 - 128$ \\
IMDb&\,SparseVD  & {\bf 85.1} & 1135x & 4611 & $16 - 17$ \\
%\multicolumn{1}{r}{\textcolor{gray}{(ours)}\!\!}
&\,SparseVD-Voc  & 83.6 & {\bf 18792x} & {\bf 292} & $\bf 1 - 8$ \\ 
\hline
&\,Original  & {\bf 90.6} & 1x & 20000 & $300 - 512$ \\
AGNews&\,SparseVD  & 88.8 & 322x & 5727 & $179 - 56$\\
&\,SparseVD-Voc  & 89.2 & {\bf 469x} & {\bf 2444} & $\bf 127-32$\\ 
\end{tabular}
\caption{Results on text classification tasks. Compression is equal to $|w|/|w\neq0|$. In last two columns number of remaining neurons in the input, embedding and recurrent layers are reported.}\label{tab:class}
\end{table*}

\begin{table*}[ht!]
  \normalsize
  \centering
  \begin{tabular}{clccccc}
Task & Method & Valid & Test & Compression  & Vocabulary & Neurons $h$\\ 
\hline
&\,Original  & 1.499 & 1.454 & 1x & 50 & 1000 \\
Char PTB&\,SparseVD  & 1.472 & 1.429 & {\bf 7.9x}  & 50 & {\bf 431} \\%{\bf 4.2x}
\multicolumn{1}{c}{\textcolor{gray}{Bits-per-char}\!\!}
&\,SparseVD-Voc  & {\bf 1.458} & {\bf 1.417} & 6.0x & {\bf 46} & 510 \\ %3.53x
\hline
&\,Original  & 135.6 &	129.3 & 1x & 10000 & 256 \\
Word PTB&\,SparseVD  & {\bf 115.0} & {\bf 109.2} & {\bf 22.1x}  & 9990 & {\bf 156}  \\
\multicolumn{1}{c}{\textcolor{gray}{Perplexity}\!\!}
&\,SparseVD-Voc  & 126.0 & 120.2 & 19.3x & {\bf 3164} & 209 \\ 
\end{tabular}
\caption{Results on language modeling tasks. 
  Compression is equal to $|w|/|w\neq0|$. In last two columns number of remaining neurons in input and recurrent layers are reported.}\label{tab:lm}
\end{table*}

\subsection{Text Classification}

We evaluated our approach on two standard datasets for text classification: IMDb dataset~\cite{IMDB} for binary classification and AGNews dataset~\cite{agnews} for four-class classification. We set aside 15\% and 5\% of training data for validation purposes respectively. For both datasets, we use the vocabulary of 20,000 most frequent words.

We use networks with one embedding layer of 300 units, one LSTM layer of 128 / 512 hidden units for IMDb / AGNews, and finally, a fully connected layer applied to the last output of the LSTM. Embedding layer is initialized with word2vec~\cite{NIPS2013_5021} / GloVe~\cite{pennington2014glove} and SparseVD and SparseVD-Voc models are trained for 800 / 150 epochs on IMDb / AGNews. 

\begin{figure}[h]
    \centering
        \centering
                \includegraphics[width=\linewidth]{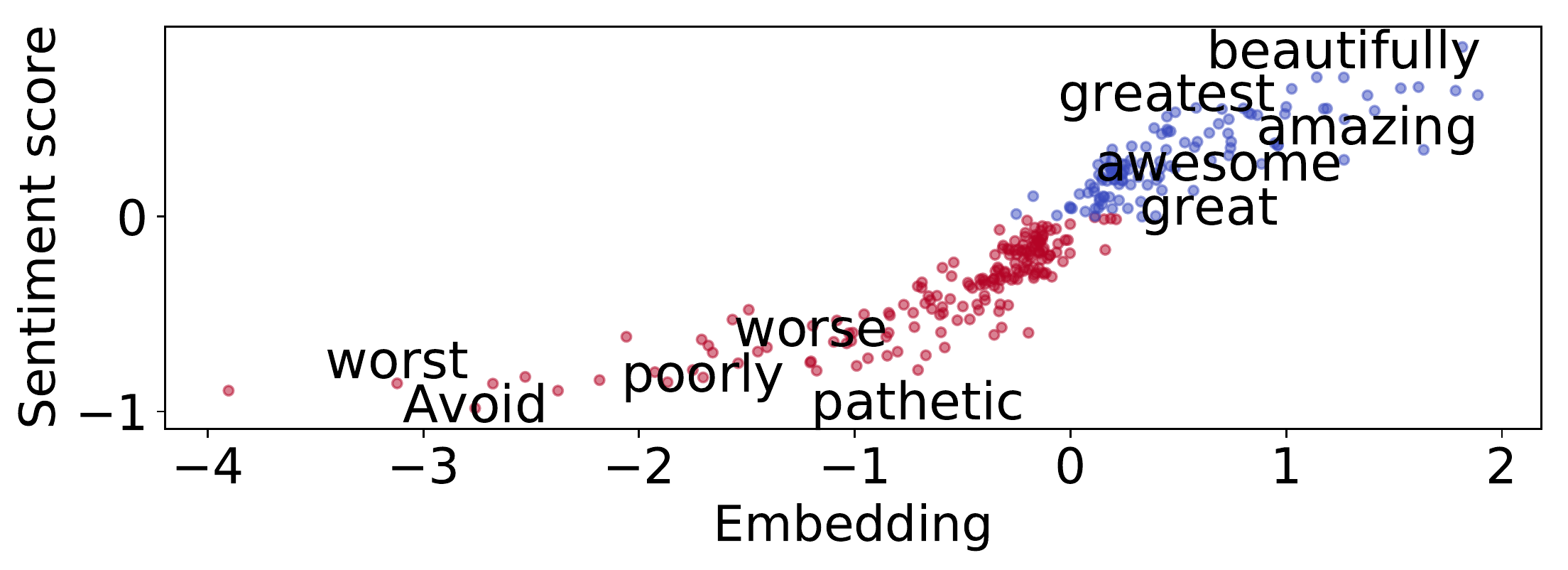}
        \caption{IMDB: remaining embedding component vs sentiment score ((\#pos. - \#neg.)\,/\,\#all texts with the word).}
        \label{fig:words}
\end{figure}

The results are shown in Table~\ref{tab:class}. SparseVD leads to a very high compression rate without a significant quality drop. SparseVD-Voc boosts the compression rate even further while still preserving the accuracy. Such high compression rates are achieved mostly because of the sparsification of the vocabulary: to classify texts we need to read only some important words from them. The remaining words in our models are mostly interpretable for the task (see Appendix~\ref{imdb} for the list of remaining words for IMBb). 
Figure~\ref{fig:words} shows the only kept embedding component for remaining words on IMDb. This component reflects the sentiment score of the words.

\subsection{Language Modeling}
We evaluate our models on the task of character-level and word-level language modeling on the Penn Treebank corpus~\cite{ptb} according to the train/valid/test partition of \citeauthor{mikolov11}~\shortcite{mikolov11}. 
The dataset has a vocabulary of 50 characters or 10,000 words.

To solve character / word-level tasks we use networks with one LSTM layer of 1000 / 256 hidden units and fully-connected layer with softmax activation to predict next character or word. We train SparseVD and SparseVD-Voc models for 250 / 150 epochs on character-level / word-level tasks. 

The results are shown in Table~\ref{tab:lm}. To obtain these results we employ LRT on the last fully-connected layer. In our experiments with language modeling LRT on the last layer accelerate the training without harming the final result. Here we do not get such extreme compression rates as in the previous experiment but still, we are able to compress the models several times while achieving better quality w.r.t. the baseline because of the regularization effect of SparseVD. Vocabulary is not sparsified in the character-level task because there are only 50 characters and all of them matter. In the word-level task more than a half of the words are dropped. However, since in language modeling almost all words are important, the sparsification of the vocabulary makes the task more difficult to the network and leads to the drop in quality and the overall compression (network needs more difficult dynamic in the recurrent layer).

\section*{Acknowledgments}
 Results on SparseVD for RNNs shown in Section 3.2 have been supported by Russian Science Foundation (grant 17-71-20072). Results on  multiplicative weights for vocabulary sparsification shown in Section 3.3 have been supported by Samsung Research, Samsung Electronics.
 \\

\bibliographystyle{acl_natbib_nourl}
\bibliography{emnlp2018}
\clearpage
\newpage
    
\appendix
\section{Experimental setup} \label{setup}

{\bf Initialization for text classification.} Hidden-to-hidden weight matrices $W^h$ are initialized orthogonally and all other matrices are initialized uniformly using the method by~\citeauthor{pmlr-v9-glorot10a}~\shortcite{pmlr-v9-glorot10a}. 

We train our networks using batches of size 128 and a learning rate of 0.0005. 

{\bf Initialization for language modeling.}
All weight matrices of the networks are initialized orthogonally and all biases are initialized with zeros. Initial values of hidden and cell elements are not trainable and equal to zero.

For the character-level task, we train our networks on non-overlapping sequences of 100 characters in mini-batches of 64 using a learning rate of 0.002 and clip gradients with threshold 1. 

For the word-level task, networks are unrolled for 35 steps. We use the final hidden states of the current mini-batch as the initial hidden state of the subsequent mini-batch (successive mini batches sequentially traverse the training set). The size of each mini-batch is 32. We train models using a learning rate of 0.002 and clip gradients with threshold 10.

\section{A list of remaining words on IMDB}  \label{imdb}
SparseVD with multiplicative weights retained the following words on IMDB task (sorted by descending frequency in the whole corpus):

start, oov, and, to, is, br, in, it, this, was, film, t, you, not, have, It, just, good, very, would, story, if, only, see, even, no, were, my, much, well, bad, will, great, first, most, make, also, could, too, any, then, seen, plot, acting, life, over, off, did, love, best, better, i, If, still, man, something, m, re, thing, years, old, makes, director, nothing, seems, pretty, enough, own, original, world, series, young, us, right, always, isn, least, interesting, bit, both, script, minutes, making, 2, performance, might, far, anything, guy, She, am, away, woman, fun, played, worst, trying, looks, especially, book, DVD, reason, money, actor, shows, job, 1, someone, true, wife, beautiful, left, idea, half, excellent, 3, nice, fan, let, rest, poor, low, try, classic, production, boring, wrong, enjoy, mean, No, instead, awful, stupid, remember, wonderful, often, become, terrible, others, dialogue, perfect, liked, supposed, entertaining, waste, His, problem, Then, worse, definitely, 4, seemed, lives, example, care, loved, Why, tries, guess, genre, history, enjoyed, heart, amazing, starts, town, favorite, car, today, decent, brilliant, horrible, slow, kill, attempt, lack, interest, strong, chance, wouldn, sometimes, except, looked, crap, highly, wonder, annoying, Oh, simple, reality, gore, ridiculous, hilarious, talking, female, episodes, body, saying, running, save, disappointed, 7, 8, OK, word, thriller, Jack, silly, cheap, Oscar, predictable, enjoyable, moving, Unfortunately, surprised, release, effort, 9, none, dull, bunch, comments, realistic, fantastic, weak, atmosphere, apparently, premise, greatest, believable, lame, poorly, NOT, superb, badly, mess, perfectly, unique, joke, fails, masterpiece, sorry, nudity, flat, Good, dumb, Great, D, wasted, unless, bored, Tony, language, incredible, pointless, avoid, trash, failed, fake, Very, Stewart, awesome, garbage, pathetic, genius, glad, neither, laughable, beautifully, excuse, disappointing, disappointment, outstanding, stunning, noir, lacks, gem, F, redeeming, thin, absurd, Jesus, blame, rubbish, unfunny, Avoid, irritating, dreadful, skip, racist, Highly, MST3K

An interesting observation is that low and high ratings of films are kept while the middle values (5, 6) of the ratings are dropped.

\end{document}